\documentclass[10pt,twocolumn,letterpaper]{article}

\usepackage{cvpr}
\usepackage{times}
\usepackage{epsfig}
\usepackage{graphicx}
\usepackage{amsmath}
\usepackage{amssymb}

\usepackage[breaklinks=true,bookmarks=false]{hyperref}

\cvprfinalcopy 

\def\cvprPaperID{****} 

\begin{document}

\title{6D Object Pose Estimation without PnP}

\author{Jin Liu$^1$$^,$$^2$\\
$^1$Wuhan University\\
$^2$Wuhan Xiongchugaojing Tech Company\\
Wuhan, China\\
{\tt\small jliu@sgg.whu.edu.cn}
\and
Sheng He$^1$\\
$^1$Wuhan University\\
Wuhan, China\\
{\tt\small whusheng1996@163.com}
}

\maketitle

\begin{abstract}
   In this paper, we propose an efficient end-to-end algorithm
to tackle the problem of estimating the 6D pose of
objects from a single RGB image. Our system trains a fully
convolutional network to regress the 3D rotation and the 3D
translation in region layer. On this basis, a special layer,
Collinear Equation Layer, is added next to region layer to
output the 2D projections of the 3D bounding boxs corners.
In the back propagation stage, the 6D pose network are adjusted
according to the error of the 2D projections. In the
detection phase, we directly output the position and pose
through the region layer. Besides, we introduce a novel and
concise representation of 3D rotation to make the regression
more precise and easier. Experiments show that our method
outperforms base-line and state of the art methods both at
accuracy and efficiency. In the LineMod dataset, our algorithm
achieves less than 18 ms/object on a GeForce GTX
1080Ti GPU, while the translational error and rotational
error are less than 1.67 cm and 2.5 degree.
\end{abstract}

\section{Introduction}

Object detection and localization has always been a
hot topic of computer vision. Traditional methods, like
YOLO[1], SSD[2] and Mask R-CNN[3], have experienced
a tremendous success in 2D domain. However, those methods
cant achieve accurate semantic understanding of the objective
three-dimensional world. The ultimate goal of computer
vision is to study the nature of the objective threedimensional
world through images. To tackle this, more
and more attention has been paid to object 6D pose estimation.
Real-time 6D pose estimation is crucial for augmented
reality, virtual reality and robotics.

Currently, feature-based methods[4,5,6], template-based
methods[7,8] and RGB-D methods[9,10,11,12,13] have
achieved robust results to some extent. Feature-based methods
tackled this task by matching feature points between
3D models and images. However, only when there are
rich textures on the objects that those methods work. As
a result, they are unable to handle texture-less objects[14].
Template-based methods use a rigid template to match different
locations in the input image. Such methods are likely
to be affected by occlusions. RGB-D methods use depth
data as additional information, which simplifies the task.
However, active depth sensors are power hungry, which
makes 6D objective detection methods for passive RGB images
more attractive for mobile and wearable cameras[15].
Besides, acquiring depth data needs additional hardware
costs.

Deep learning techniques have recently become mainstream
to estimate 6D object pose, among which [15] and
[21] are two typical examples. In [15, 21], CNNs are used to
predict 2D projections of 3D bounding boxs corners (for the
sake of simplification, we call the 2D projections pts), then
6D pose are obtained by PnP algorithm. The deficiency of
the two methods is that PnP costs extra time, decreasing the
efficiency. In this paper, we propose a generic framework
which overcomes the shortcomings of existing methods to
estimate 6D object pose. We extend YOLO V2[26] to perform
6D pose estimation from single RGB images. In the
training phase, we feed images to the fully convolutional
channels to output the 3D translation parameters (tu, tv, tw)
and 3D rotation parameters (a, b, c). The special layer,
Collinear Equation Layer, follows the meta-architecture of
YOLO with architecture adaptation and tuning to predict
the pts. Then we adjust the network with the pts error. Unlike
[15] and [21], in the testing phase, we discard Collinear
Equation Layer and directly predict 6D pose parameters.

Our work has the following advantages and contributions:

a) We propose a novel method for 6D pose estimation in
a really end-to-end manner. We bring in Collinear Equation
Layer to regress 2D projections of 3D bounding boxs corners
to train our network. In the testing stage, we discard
the last layer and directly obtain 6D pose, avoiding PnP algorithm,
which makes the estimation fast and accurate.

b) We introduce a brand new representation for 3D rotation.
This representation is easy to regression.

c) Extending YOLO V2 to directly predict 6D pose.

The remainder of the paper is organized as follows. After
the overview of related work, we introduce our method.Then 
we display the experimental results, followed by the
final conclusion.

\section{Related Work}

The literature on 6D pose estimation is very large
and we have mentioned some in the previous section,
thus we will focus only on recent works based on deep
learning. Most of recent works use CNN to solve 6D
pose problems, including camera pose[16, 17] and object
pose[15,18,19,20,21,22,23,24].

In [16,17], the authors train CNNs to directly regress 6D
camera pose from a single RGB image. The camera pose
estimation is much easier than the object pose estimation,
for there is no need to detect any object.

In [18,19], the authors use CNNs to regress 3D object
pose directly, their works focus only on 3D rotation estimation
while 3D translation is not included. In [20], SSD detection
framework is extended to 6D pose estimation. The
authors transform pose detection into two-stage classification
tasks, view angle classification and in-plane rotation
classification. However, wrong classification in either stage
could cause an incorrect pose estimation. In [22], the authors
first use a CNN to regress 3D object orientation, then
combines these estimates with geometric constraints provided
by a 2D object bounding box to produce a complete
3D bounding box. However, in general, this method needs
to solve 4096 linear equations. In special circumstances,
such as the KITTI dataset[25], object pitch and roll angles
are both zero, there are still 64 equations to be solve, which
makes the method computational costly. In [23], the pose
parameters are decoupled into translation and rotation, then
the rotation is regressed via a Lie algebra representation.
This method assumes that the 2D projection of the objects
origin is in the 2D boxs center, which makes the estimation
of translation inaccuracy. In [24], a feedback loop consisting
of deep networks are developed for 6D pose estimation.
In this method, the inaccurate pose data is re-projected and
compared with the original image for accurate correction.
However, the preparation of sample is intricate.

BB8[21] provides a precise method to estimation 6D
pose. the authors firstly use a segmentation network to localize
objects. Then another CNN is used to predict the 2D
projections of the 3D bounding boxs corners around the object.
The 6D pose is estimated through a PnP algorithm.
Finally, a CNN is trained to refine the pose. The method
is multi-stage, which increases their running time. Similar
to BB8, [15] detects the 2D projections of the corners,
too. But the authors use a direct way by propose a singleshot
deep CNN architecture, then employ PnP algorithm to
get the 6D pose. Both [21] and [15] achieve high accuracy.
However, the two method employ PnP algorithm to attain
6D pose, which is not really end-to-end, and the PnP algorithm
will cost extra time. Besides, the regression of each
corner is independent and no constraint exists. This may result
in that some corners are inaccurately predicted, which
will have bad impact on the PnP algorithm. Compared to
them, our method regresses the corners with constraint produced
by Collinearity Equation Layer in the training stage,
but directly predict 6D pose while testing. In this way, we
avoid the shortcomings raised by PnP.

\section{Method}
\subsection{Position parameter}
According to the small hole imaging equation we have the following formula:

\begin{equation}
\begin{split}
{{t}} &= {{z}}{K^{ - 1}}\left[ {\begin{array}{*{20}{c}}
u\\
v\\
1
\end{array}} \right] \\ &= {\rm{z}}{K^{ - 1}}\left[ {\begin{array}{*{20}{c}}
{W({c_0} + 0.5 + \Delta u)/w}\\
{H\left( {{r_0} + 0.5 + \Delta v} \right)/h}\\
1
\end{array}} \right]
\end{split}
\end{equation}

considering

\begin{equation}
{K^{ - 1}} = \left[ {\begin{array}{*{20}{c}}
{\frac{1}{{{f_x}}}}&0&{ - \frac{{{c_x}}}{{{f_x}}}}\\
0&{\frac{1}{{{f_y}}}}&{ - \frac{{{c_y}}}{{{f_y}}}}\\
0&0&1
\end{array}} \right]
\end{equation}

we get

\begin{equation}
{\bf{t}} = {\bf{b}}{{\bf{e}}^{tw}}\left[ {\begin{array}{*{20}{c}}
{\frac{1}{{{{\bf{f}}_x}}}\left( {\frac{{W({c_0} + 0.5 + s\sigma \left( {{t_u}} \right))}}{w} - {c_x}} \right)}\\
{\frac{1}{{{{\bf{f}}_y}}}\left( {\frac{{H({r_0} + 0.5 + s\sigma \left( {{t_v}} \right))}}{h} - {c_y}} \right)}\\
1
\end{array}} \right]
\end{equation}

where $\sigma(.)$ is the loggy excitation function and s is an adjustable parameter. Considering that the object may be distributed over a large range, we take s=4.0. The neural network outputs three translation variables tu, tv, tw $\rightarrow$ [$\Delta$u, $\Delta$v, z]$\rightarrow$[X Y Z]=t.

\subsection{Pose parameter}
The rotation matrix R in the collinear equation can perfectly express the rotation of the camera relative to the object. However, the R matrix is not suitable for direct prediction using neural networks. Because R is a unit orthogonal matrix, there are too many redundancy, so we use abc conversion:

\begin{equation}
\begin{split}
R &= \frac{1}{{1 + {a^2} + {b^2} + {c^2}}}\\&\left[\!\!\!\! {\begin{array}{*{20}{c}}
{1 + {a^2} - {b^2} - {c^2}}&{2ab - 2c}&{2ac + 2b}\\
{2ab + 2c}&{1 - {a^2} + {b^2} - {c^2}}&{2bc - 2a}\\
{2ac - 2b}&{2bc + 2a}&{1 - {a^2} - {b^2} + {c^2}}
\end{array}} \!\!\!\!\right]
\end{split}
\end{equation}

The abc can be predicted by the neural network and then the (4) equation can be used to obtain the pose matrix. The abc transform does not need to worry about the angle loop problem, and there is no redundancy without worrying about the unitized constraint problem. Therefore, the abc is selected for network pose prediction.

\subsection{Overall pipeline}

The main idea of this paper is to propose a full convolution network that implements 6DPose. This idea is to add a collinear equation layer after the region layer of the deep network. In the training phase, the region layer predicts the position and rotation parameters R and t. The coordinates u, v are backpropagated by regression 2D pts to correct R and t. In the prediction phase, R and t are directly obtained by the region layer. Figure 1 shows the pipeline.

\begin{figure*}
\begin{center}
\centerline{\includegraphics[width=16cm]{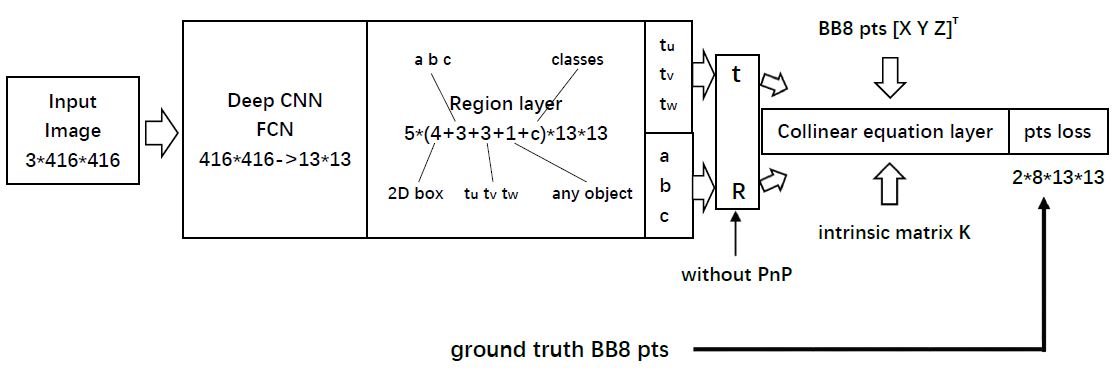}}
\end{center}
   \caption{Overall pipeline.}
\end{figure*}
The input of the neural network is a 3*416*416 color image, which are converted into a 13*13 array through a full convolutional network, and output 5*(4+3+3+1+c) channels through a 13x13 arrays region layer, in which 4 channels record 2D box coordinate information, three channels abc record rotation data, three channels tu,tv,tw are responsible for 3d translation. one channel is responsible for whether the object is near the cell, c channel softmax outputs the target category.

In order to improve the detection accuracy, we have designed five anchors with reference to yolov2, which are used to extract objects of different scales.
\begin{figure*}
\begin{center}
\centerline{\includegraphics[width=12cm]{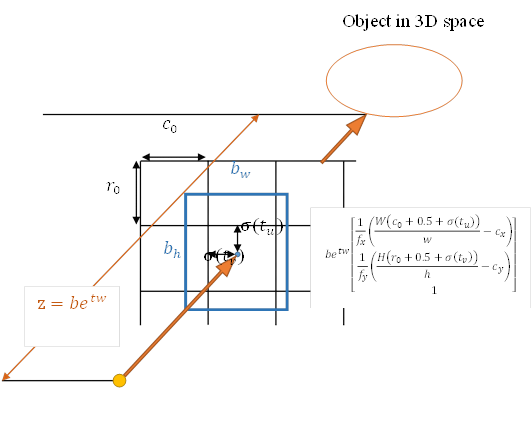}}
\end{center}
   \caption{Schematic diagram of the 3D position prediction of the region layer.}
\end{figure*}

\section{Collinear Equation Layer}
As shown in Fig.1, The Collinear Equation Layer is responsible for adjusting the position and rotation parameters by pts error back propagation.

\subsection{Forward propagation}

\begin{equation}
z\left[ {\begin{array}{*{20}{c}}
u\\
v\\
1
\end{array}} \right] = KR\left( {X - T} \right) = K\left( {RX + t} \right)
\end{equation}

The last layer in Figure 1 is a mapping operation that implements small hole imaging. Divide the first line of equation (5) by the third line, and divide the second line by the third line to get the collinear equation as follows:

\begin{equation}
\left\{ \begin{array}{l}
u = {c_x} + {f_x}\frac{X}{Z} = {c_x} + {f_x}\frac{{{r_{11}}{X_x} + {r_{12}}{X_y} + {r_{13}}{X_z} + {{\rm{t}}_x}}}{{{r_{31}}{X_x} + {r_{32}}{X_y} + {r_{33}}{X_z} + {t_z}}}\\
v = {c_y} + {f_y}\frac{Y}{Z} = {c_y} + {f_y}\frac{{{r_{21}}{X_x} + {r_{22}}{X_y} + {r_{23}}{X_z} + {t_y}}}{{{r_{31}}{X_x} + {r_{32}}{X_y} + {r_{33}}{X_z} + {t_z}}}
\end{array} \right\}
\end{equation}

Among them, $\text{r}_{11}$-$\text{r}_{33}$ are the 9 elements of the unit orthogonal matrix R, which can be expressed by abc in Eq.(4).The formula (6) can be regarded as the forward propagation formula of the collinear equation layer.
Where R is obtained by abc output from the neural network region layer through the formula (4), and t is obtained by the formula (3).

\begin{figure}[bh]
\centerline{\includegraphics[width=8cm]{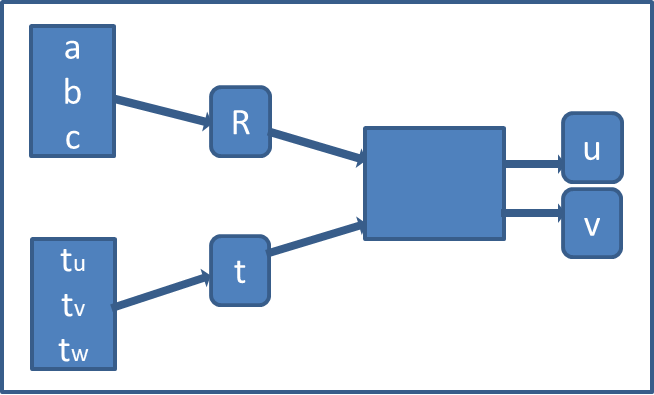}}
\vspace*{8pt}
\caption{Forward propagation calculation process}
\end{figure}

\subsection{Backward propagation}
The collinear equation layer is only used during the training phase in order to pass the error of the pts to the 6D pose parameters in region layer error, so that the error weights of the R and t parameters can be reasonably allocated. Defined according to the definition of error:

\begin{equation}
loss{\rm{ }} = {\rm{ }}{\rm{|pt}}{{\rm{s}}_{predict}} - (\overline {pts} ){|_2}
\end{equation}

\begin{equation}
\frac{{\partial E_{qt}^2}}{{\partial u}} = \mathop \sum \limits_{i = 1}^n {\left( {{\rm{ui}} - \overline {{\rm{ui}}} } \right)^2}
\end{equation}

\begin{equation}
\frac{{\partial E_{qt}^2}}{{\partial v}} = \mathop \sum \limits_{i = 1}^n {\left( {{\rm{vi}} - \overline {{\rm{vi}}} } \right)^2}
\end{equation}

Where n is the number of pts points. The figure below shows the backward propagation process.
\begin{figure}[bh]
\centerline{\includegraphics[width=8cm]{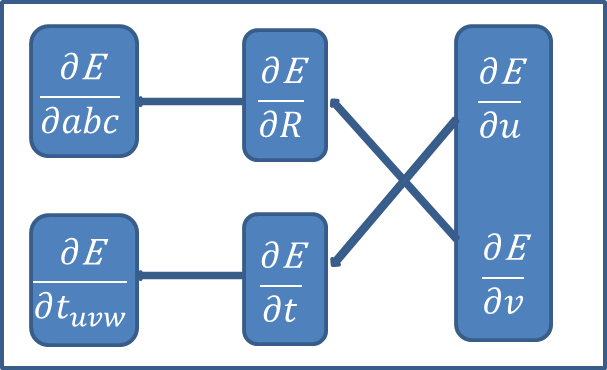}}
\vspace*{8pt}
\caption{Collinear equation layer backward propagation process}
\end{figure}

\section{Overall network structure}
\subsection{Network Design}
In Figure 1, DeepCNN is a mapping from 3*416*416 to 13*13*5*(4+3+3+1+c). We designed two fully convolutional networks for 6DPose prediction. Network 6D Pose-linemod-13c for 3D mesh ply dataset for LineMod[7]. Network 6D Pose-voc-8c for 8 typical objects in the VOC2007+VOC2012 dataset.

In network 6DPose-linemode-13c, we design a c=13 category object for linemode 3DMesh, conv layer29 output 5*(11+13)=120 channels.
\begin{table}
\begin{center}
\begin{tabular}{|l|c|l|c|l|}
\hline
layer & filters & size & input & output\\
\hline\hline
0 conv & 32 & 3*3/1 & 416*416*3 & 416*416*32\\
1 max &  & 2*2/2 & 416*416*32 & 208*208*32\\
2 conv & 64 & 3*3/1 & 208*208*32 & 208*208*64\\
3 max & & 2*2/2 & 208*208*64 & 104*104*64\\
4 conv & 128 & 3*3/1 & 104*104*64 & 104*104*128\\
5 conv & 64 & 1*1/1 & 104*104*128 & 104*104*64\\
6 conv & 128 & 3*3/1 & 104*104*64 & 104*104*128\\
7 max & & 2*2/2 & 104*104*128 & 52*52*128\\
8 conv & 256 & 3*3/1 & 52*52*128 & 52*52*256\\
9 conv & 128 & 1*1/1 & 52*52*256 & 52*52*128\\
10 conv & 256 & 3*3/1 & 52*52*128 & 52*52*256\\
11 max & & 2*2/2 & 52*52*256 & 26*26*256\\
12 conv & 512 & 3*3/1 & 26*26*256 & 26*26*512\\
13 conv & 256 & 1*1/1 & 26*26*512 & 26*26*256\\
14 conv & 512 & 3*3/1 & 26*26*256 & 26*26*512\\
15 conv & 256 & 1*1/1 & 26*26*512 & 26*26*256\\
16 conv & 512 & 3*3/1 & 26*26*256 & 26*26*512\\
17 max & & 2*2/2 & 26*26*512 & 13*13*512\\
18 conv & 1024 & 3*3/1 & 13*13*512 & 13*13*1024\\
19 conv & 512 & 1*1/1 & 13*13*1024 & 13*13*512\\
20 conv & 1024 & 3*3/1 & 13*13*512 & 13*13*1024\\
21 conv & 512 & 1*1/1 & 13*13*1024 & 13*13*512\\
22 conv & 1024 & 3*3/1 & 13*13*512 & 13*13*1024\\
23 conv & 1024 & 3*3/1 & 13*13*1024 & 13*13*1024\\
24 conv & 1024 & 3*3/1 & 13*13*1024 & 13*13*1024\\
25 route & 16 & & & \\
26 reorg & &  /2 & 26*26*512 & 13*13*2048\\
27 route & 26 & 24 & &\\
28 conv & 1024 & 3*3/1 & 13*13*3072 & 13*13*1024\\
29 conv & 220 & 1*1/1 & 13*13*1024 & 13*13*120\\
30 region & & & & 13*13*5*(4+3+3+1+13)\\
\hline
\end{tabular}
\end{center}
\caption{Structure of network 6DPose-linemode-13c}
\end{table}

\subsection{3D mesh sample training and augmentation}
3D augmentation is implemented using the document and the OpenGL rendering algorithm described in [20].

\subsection{VOC 2D image sample training and augmentation}
We have built a VOC3D dataset. In order to improve the labeling speed, we use the mouse to pull out the XYZ three axes on the image to determine the target's posture R={r11,...,r33}. Suppose the user uses the mouse to mark the axis vector dx, dy, dz direction of the object on the image, and the linear equation au+bv+c=0 on the corresponding image satisfies the equation below:

\begin{equation}
\left[ {\begin{array}{*{20}{c}}
a&b&c
\end{array}} \right]KR\left[ {\begin{array}{*{20}{c}}
{{d_x}}\\
{{d_y}}\\
{{d_z}}
\end{array}} \right] = 0
\end{equation}

That is, the rotation R is the solution of equation (10). The rotation data R can be solved by using the LM algorithm. Translation T is the solution to the equation below:

\begin{equation}
\begin{split}
&\left[ {\tiny\begin{array}{*{20}{c}}
{{u_{KiL}}{r_{31}} - {r_{11}}}&{{u_{KiL}}{r_{32}} - {r_{12}}}&{{u_{KiL}}{r_{33}} - {r_{13}}}\\
{{u_{KiR}}{r_{31}} - {r_{11}}}&{{u_{KiR}}{r_{32}} - {r_{12}}}&{{u_{KiR}}{r_{33}} - {r_{13}}}\\
{{v_{KiT}}{r_{31}} - {r_{21}}}&{{v_{KiT}}{r_{32}} - {r_{22}}}&{{v_{KiT}}{r_{33}} - {r_{23}}}\\
{{v_{KiB}}{r_{31}} - {r_{21}}}&{{u_{KiB}}{r_{32}} - {r_{22}}}&{{v_{KiB}}{r_{33}} - {r_{23}}}
\end{array}} \right]{\small\rm{T}} \!\!=\!\!\\\
&\begin{bmatrix}\!\!\!\!\!\!\!\!\begin{smallmatrix} \begin{smallmatrix}\begin{smallmatrix} \begin{smallmatrix}{\tiny\begin{array}{*{20}{c}}
{\left( {{u_{KiL}}{r_{31}} - {r_{11}}} \right){x_{iL}} + \left( {{u_{KiL}}{r_{32}} - {r_{12}}} \right){y_{iL}} + \left( {{u_{KiL}}{r_{33}} - {r_{13}}} \right){z_{iL}}}\\
{\left( {{u_{KiR}}{r_{31}} - {r_{11}}} \right){x_{iR}} + \left( {{u_{KiR}}{r_{32}} - {r_{12}}} \right){y_{iR}} + \left( {{u_{KiR}}{r_{33}} - {r_{13}}} \right){z_{iR}}}\\
{\left( {{v_{KiT}}{r_{31}} - {r_{21}}} \right){x_{iT}} + \left( {{u_{KiT}}{r_{32}} - {r_{22}}} \right){y_{iT}} + \left( {{u_{KiT}}{r_{33}} - {r_{23}}} \right){z_{iT}}}\\
{\left( {{v_{KiB}}{r_{31}} - {r_{21}}} \right){x_{iB}} + \left( {{u_{KiB}}{r_{32}} - {r_{22}}} \right){y_{iB}} + \left( {{u_{KiB}}{r_{33}} - {r_{23}}} \right){z_{iB}}}
\end{array}} \end{smallmatrix}\end{smallmatrix}\end{smallmatrix}\end{smallmatrix}\!\!\!\!\!\!\!\!\end{bmatrix}\!\!\!\!
\end{split}
\end{equation}

Thus, we can obtain the displacement data t of the labeled data only by solving only one linear equation. According to the pose data and the three values of the length, width and height of the target, 8 virtual feature point coordinates of the target can be obtained as the training data of our algorithm. The 2DImage data augmentation uses the affine transformation to perform a proportional translation, scaling and rotation transformation of the image and the virtual feature point coordinate synchronization.

We selected 8 types of objects suitable for 6DPose from 20 categories of objects in VOC2007 and VOC2012 to create a small VOC3d data set as follows: aero plane, boat, bus, car, chair, motorbike, person, tv monitor.

According to the discussion in Sec.5.1, the output of the last layer of the fully convolutional network is 13*13, 5*(11+c) channel data, where c=8, and the region layer accesses 95 from the 29 conv layer.

From the VOC tag image, 500 image samples were taken for testing, and the remaining 6538 images were used for training. 6D Pose prediction is shown in next section.

\section{Experiments}
As described in Sec.5.1, we constructed two networks to train 3D mesh file recognition for linemode and 2D image data for VOC2007 and VOC2012, and then predict 6D Pose for the trained model. We then evaluated the accuracy and speed of 6DPose prediction.

\subsection{Evaluation for LineMod dataset}
Firstly, we use the object.xyz file in linemode to train the linemode-6DPose-c13 model. The linemode gives the rot file and the tra file corresponding to the target's pose and translation GroundTruth values. The 6DPose prediction projection cube and the GroundTruth projection cube are overlapping displayed as follows:
\begin{figure*}
\begin{center}
\centerline{\includegraphics[width=12cm]{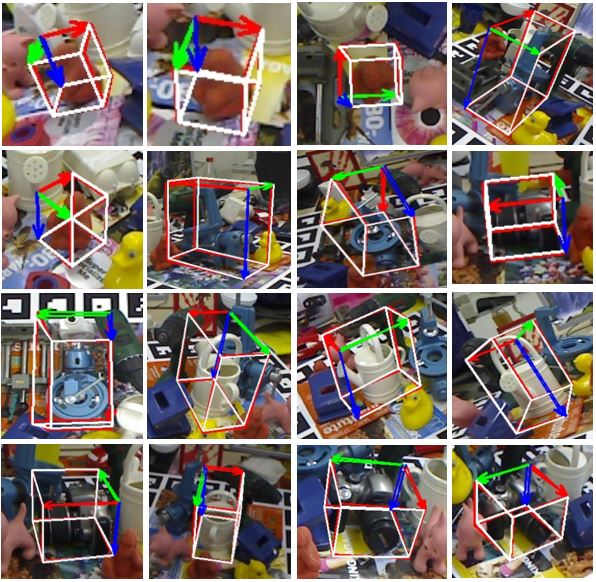}}
\end{center}
   \caption{LineMod object 6D pose prediction demo, overlapping displayed.}
\end{figure*}

The white box in Figure 5 is the GroundTruth object cube, and the red+green+blue cube is the projection corresponding to the network predicted R-t, and the two Cubes can basically overlap. The direction of the red arrow is the positive X-axis of the object itself, the direction of the green arrow is the positive direction of the Y-axis, and the direction of the blue arrow is the positive direction of the Z-axis. To evaluate our method and compare it with state of the art method, we use two metrics, 2D projection error[30] and rotational and translational error[31].

\begin{table*}[htbp]
\begin{center}
\begin{tabular}{|l|c|l|c|l|c|}
\hline
Object & Average pixel projection error & 5 pixels accuracy & 5 pixels accuracy & 5 pixels accuracy\\
\hline\hline
    & Our method & Ours & SS6D[27] & BB8[21]\\
ape & 1.98 & 0.9894 & 0.9210 & 0.9530\\
cam & 2.64 & 0.9658 & 0.9324 & 0.809\\
glue & 2.67 & 0.9680 & 0.9653 & 0.890\\
box & 2.54 & 0.9457 & 0.9033 & 0.879\\
can & 3.17 & 0.9130 & 0.9744 & 0.841\\
lamp & 2.50 & 0.9347 & 0.7687 & 0.744\\
bench & 4.25 & 0.7152 & 0.9506 & 0.800\\
cat & 2.53 & 0.9826 & 0.9741 & 0.970\\
hole & 2.61 & 0.9352 & 0.9286 & 0.905\\
duck & 2.58 & 0.9534 & 0.9465 & 0.812\\
iron & 2.52 & 0.9015 & 0.8294 & 0.789\\
driller & 2.60 & 0.8985 & 0.7941 & 0.7941\\
phone & 2.69 & 0.9458 & 0.8607 & 0.776\\
\textbf{average} & 2.71 & 0.9268 & 0.9037 & 0.839\\
bowl & 2.67 & 0.9562 &  & \\
cup & 2.98 & 0.9325 &  & \\
\hline
\end{tabular}
\end{center}
\caption{State of the art comparison of our method against SS6D and BB8 using 2D projection error.As in [28], we use the percentage of correctly predicted poses for each object. A pose is considered correct if the 2D projection pixel error is less than 5 pixels. The second column is the average pixel porjection error of our method. The last three columns are the comparison of our method against the state of art methods in 5 pixels manner.}
\end{table*}

\begin{table}
\begin{center}
\begin{tabular}{|l|c|l|c|l|c|}
\hline
Object & ours($\text{e}_{TE}$) & BB8($\text{e}_{TE}$) & ours($\text{e}_{RE}$) & BB8($\text{e}_{RE}$)\\
\hline\hline
ape & 1.88 & 1.85 & 2.45 & 2.54\\
cam & 1.85 & 1.89 & 2.43 & 2.55\\
glue & 1.67 & 1.98 & 2.51 & 2.38\\
box & 1.54 & 1.78 & 2.10 & 2.40\\
can & 1.80 & 1.97 & 2.38 & 2.13\\
lamp & 1.50 & 1.67 & 2.26 & 2.10\\
bench & 1.82 & 1.78 & 3.03 & 2.83\\
cat & 1.53 & 1.56 & 2.25 & 2.43\\
hole & 1.61 & 1.65 & 2.31 & 2.76\\
duck & 1.58 & 1.72 & 2.65 & 2.53\\
iron & 1.52 & 1.55 & 2.59 & 2.94\\
driller & 1.60 & 1.66 & 2.34 & 2.39\\
phone & 1.69 & 1.70 & 2.29 & 2.41\\
\textbf{average} & 1.66 & 1.75 & 2.43 & 2.49\\
\hline
\end{tabular}
\end{center}
\caption{State of the art comparison of our method against BB8 using rotational and translational error. The unit of $\text{e}_{TE}$ and $\text{e}_{RE}$ are cm and degree.}
\end{table}

From the above two tables, the performance of our method is better than that of BB8 and SS6D on both 2D project  and 6D pose accuracy evaluation criteria.

Experiments have found that for the BB8 algorithm, each pts is completely independent, and the error is determined by the max pts error of pts. For our Direct 6D Pose, pts is preceded by collinear equations, and the algorithm error depends on the overall error of pts. Therefore, Direct 6DPose is very suitable for stereo vision position and rotation prediction.

\subsection{Evaluation for VOC3D 2D images}
The prediction effect of the 6D pose-voc-8c network designed by Sec.5.3 for predicting 8 typical VOC targets is shown in Figure 6. The average pixel projection error (pixels) for 6D pose-voc-8c prediction is shown in Table 4.
\begin{figure*}
\begin{center}
\centerline{\includegraphics[width=11cm]{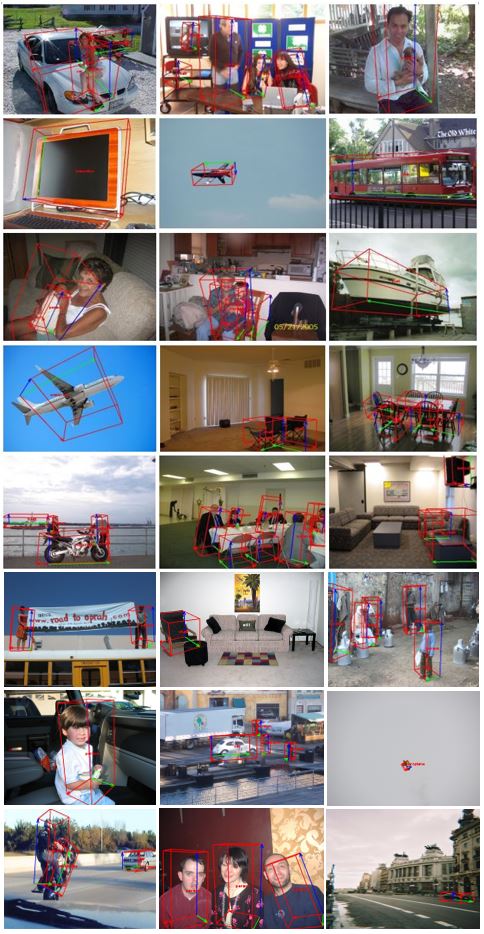}}
\end{center}
   \caption{6D pose predicted by the 6D pose-voc-c8 network.}
\end{figure*}

\begin{table}
\begin{center}
\begin{tabular}{|l|c|l|}
\hline
Object & error(8pts) & error(27pts)\\
\hline\hline
airplane & 8.38 & 6.26\\
boat & 9.89 & 8.65\\
bus & 11.10 & 7.67\\
car & 9.45 & 8.54\\
chair & 8.47 & 6.35\\
motorbike & 6.50 & 5.18\\
person & 9.25 & 8.93\\
tv monitor & 9.53 & 8.78\\
\hline
\end{tabular}
\end{center}
\caption{Average pixel projection error (pixels) for 6D pose-voc-8c prediction}
\end{table}

The cause of the error: 1. The internal parameters of the camera are not accurate; 2. Without accurate point cloud data, the length, width and height of the object are not always accurate, but if the length, width and height ratio are correct, the projected coordinates of the object on the image can still be correct.

The 6D pose R-t predicted by the network is used to draw the effect of the 3D object Cube on the image. The red arrow is the target's own X axis, and the green arrow is the target Y axis, the blue arrow is the target's Z axis. This 6Dose 2DProject error predicted by 6DPose-voc-8c is within the acceptable range of visual inspection.

\subsection{Computation times}
Our implementation takes 16-17ms for one object 6D pose prediction, on an Intel Core i7-5820K 3.30 GHz desktop with a GeForce 1080Ti. The table below shows the computation times comparsions between our method and other methods.
\begin{table}
\begin{center}
\begin{tabular}{|l|c|}
\hline
Method & computation time (ms) \\
\hline\hline
SSD-6D & 20 \\
BB8 & 130 \\
Brachmann et al. & 500\\
Rad and Lepetit & 333\\
6D pose-linemode-13c & 18\\
6D pose-voc-8c & 17\\
\hline
\end{tabular}
\end{center}
\caption{Computation time comparison.}
\end{table}

\section{Conclusion}
We designed an end-to-end 6D pose network which used the advantages of BB8 pts regression, but propagated the pts error back to the position and attitude error through the collinear equation layer, thus avoiding the post-processing pnp processing. In this way, the implementation consumption and additional errors caused by the Pnp algorithm are avoided. This algorithm can achieve 55 fps, the average 6dpose 5-pixel projection error is 0.928, the average translation error is less than 1.7cm, and the average rotation error is lessthan 2.5 degree. The algorithm does not require refinement or other post-processing post-processing. In the future, we will further extend the training data set from VOC data to COCO to achieve 6D pose prediction of large-scale 2D image data, so that 6D pose technology can be more practical for outdoor large-scale natural scenes.

\section{Acknowledgement}
This research was supported by Wuhan Xiongchugaojing Tech Company. We gratefully acknowledge the support
of Wuhan Xiongchugaojing Tech Company with the donation of the GTX 1080Ti GPU used for this research.



\end{document}